\documentclass[twoside,11pt]{article}

\usepackage{jmlr2e}

\usepackage[utf8]{inputenc} 
\usepackage[T1]{fontenc}    
\usepackage{hyperref}       
\usepackage{url}            
\usepackage{booktabs}       
\usepackage{amsfonts}       
\usepackage{nicefrac}       
\usepackage{microtype}      
\usepackage{xcolor}         
\usepackage{dsfont}
\usepackage{caption}
\usepackage{amsmath}

\DeclareMathOperator*{\argmax}{argmax}
\usepackage{bm}
\usepackage{todonotes}
\usepackage{graphicx}
\usepackage{xcolor}
\usepackage{colortbl}
\usepackage{subcaption}
\usepackage{float}
\usepackage{verbatimbox}
\usepackage[ruled,vlined]{algorithm2e}
\usepackage{booktabs}
\usepackage{multirow}
\usepackage{natbib}
\usepackage{listings}

\jmlrheading{1}{}{}{}{}{}

\firstpageno{1}

\begin{document}

\title{Information-theoretic Inducing Point Placement for High-throughput Bayesian Optimisation}
\author{\name Henry B. Moss \email henry.moss@secondmind.ai \\
\addr Secondmind, Cambridge, UK
\AND
\name Sebastian W. Ober \email swo25@cam.ac.uk \\
\addr Department of Engineering, University of Cambridge, \& Secondmind, Cambridge, UK
\AND
\name Victor  Picheny \email victor@secondmind.ai \\
\addr Secondmind, Cambridge, UK}

\maketitle

\begin{abstract}%
Sparse Gaussian Processes are a key component of high-throughput Bayesian optimisation (BO) loops --- an increasingly common setting where evaluation budgets are large and highly parallelised. 
By using representative subsets of the available data to build approximate posteriors, sparse models dramatically reduce the computational costs of surrogate modelling by relying on a small set of pseudo-observations,  the so-called \textit{inducing points}, in lieu of the full data set. However, current approaches to design inducing points are not appropriate within BO loops as they seek to reduce global uncertainty in the objective function. 
Thus, the high-fidelity modelling of promising and data-dense regions required for precise optimisation is sacrificed and computational resources are instead wasted on modelling areas of the space already known to be sub-optimal. 
Inspired by entropy-based BO methods, we propose a novel inducing point design that uses a principled information-theoretic criterion to select inducing points. By choosing inducing points to maximally reduce both global uncertainty \textbf{and} uncertainty in the maximum value of the objective function, we build surrogate models able to support high-precision high-throughput BO. 
\end{abstract}

\section{Introduction}

Countless design tasks in science, industry and machine learning can be formulated as high-throughput optimisation problems, as characterised by access to substantial evaluation budgets and an ability to make large batches of evaluations in parallel. Prominent examples include high-throughput screening within drug discovery, DNA sequencing, and experimental design pipelines, where automation allows researchers to efficiently oversee thousands of scientific experiments, field tests and simulations. In addition, these  pipelines employ an ever increasing degree of parallelisation, through cheap access to high-fidelity sensor arrays and cloud compute resources. However, such design tasks have sufficiently large and multi-modal search spaces that, even under large optimisation budgets, only a small proportion of candidate solutions can ever be evaluated, and often only approximately or under significant levels of observation noise. Consequently, most existing optimisation routines are unsuitable, as brute-force methods require too many evaluations and assume exact evaluations.

\begin{figure}%
\begin{center}
    \includegraphics[width= 0.8\textwidth]{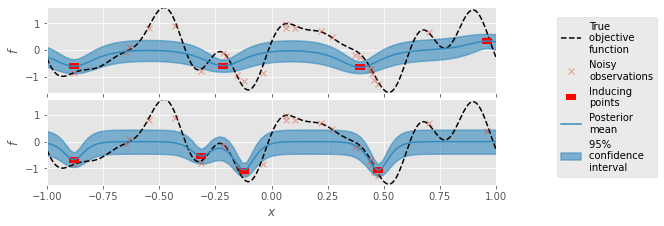}%
\end{center}
\caption{ Two sparse GP surrogate models with inducing points chosen using an existing method (top) and using our proposed BO-specific method (bottom). Our method focuses modelling resources into promising areas of the search space and provides a model more suited for directing BO.
}%
\label{fig::demo}%
\end{figure}

Bayesian Optimisation \citep[BO, see][for a review]{shahriari2016taking} has surfaced as the \textit{de facto} approach for solving black-box optimisation tasks under restricted evaluation budgets, with numerous successful applications across the empirical sciences and industry. However, vanilla BO relies on a Gaussian process \citep[GP,][]{rasmussen2003gaussian},
which incurs a significant computational overhead for each individual optimisation step, and this cost becomes increasingly unwieldy as data volumes increase,
making it unsuitable for the high-throughput tasks motivated above. 
Sparse GPs \citep{titsias2009variational} dramatically reduce the computational cost of GPs,
allowing BO to scale efficiently with data volumes \citep{vakili2021scalable}.
In a nutshell, sparse GPs replace the full set of observations by a smaller representative set of pseudo-observations referred to as \textit{inducing points}.
%
The choice of the inducing point locations has a critical influence on the behaviour of the model, in particular as it encodes its local expressivity.
However, existing approaches for inducing point placement focus purely on regression tasks, i.e. the global accuracy of models and so sacrifice the high-fidelity modelling of promising regions that is required to guide precise optimisation, leading potentially to poor optimisation performance (Figure \ref{fig::demo}).


We propose the first inducing point selection strategy designed specifically for BO.  Our subset selection method uses a principled information-theoretic criterion to explicitly trade off the modelling resources spent explaining global variations in the objective functions with those spent providing accurate modelling around potential optima (as demonstrated in Figure \ref{fig::demo}). Using synthetic benchmark problems, we demonstrate that our new inducing point selection method supports high-precision BO while existing selection strategies do not.

\section{Preliminaries}
\label{subsec::prelims}

\subsection{Bayesian Optimisation with Sparse Gaussian Processes}
GP models are a popular choice for Bayesian modelling, as they combine flexibility with reliable uncertainty estimates.
Moreover, for regression with a Gaussian likelihood, the exact posterior can be computed in closed form.
However, computing this posterior has computational complexity $O(N^3)$ and requires $O(N^2)$ memory.

To mitigate this, sparse approaches \citep{titsias2009variational, hensman2013gaussian} have been developed which have complexity $O(M^2 N)$ and $O(MN)$ memory cost, where we use $M << N$ \textit{inducing points}, which are typically selected to provide an effective representation of the full dataset.
In this work, we focus on the stochastic variational Gaussian process (SVGP) approach introduced in \citet{hensman2013gaussian}, which allows for training with minibatches.

Finally, \citet{vakili2021scalable} have recently shown that the decoupled sampling approach based on Matheron's rule introduced in \citet{wilson2020efficiently} can be used with SVGP models for BO with large batch Thompson sampling, resulting in drastic efficiency gains over traditional full GP-based BO without significant impact on regret performance.

\subsection{Inducing Point Selection Through the Lens of Experimental Design}
\label{CVR}
Common approaches for selecting inducing points include taking a random subset of the data, 
computing centroids obtained by running a k-means algorithm on the data, or directly optimising their locations by maximising the ELBO.
Recent theoretical results on sparse GP methods \citep{burt2019rates,vakili2021scalable} assume that the set of $M$ inducing points $\mathcal{Z}$ are sampled from the availible data according to an $M$-Determinantal Point Process \citep[$M$-DPP,][]{kulesza2012determinantal}. For regression tasks, this is a meaningful design criterion, as the repulsion properties of DPPs ensure that inducing points are spread across the whole search space with high probability. More precisely, an $M$-DPP inducing point selection criterion chooses the set $Z$ for our $M$ inducing points with probability proportional to the determinant of the $M\times M$ Gram matrix $K_{Z}=\left[k(\textbf{z}_i,\textbf{z}_j)\right]_{(\textbf{z}_i,\textbf{z}_j)\in Z\times Z}$, i.e.,
\begin{align}
    \mathds{P}(\mathcal{Z}=Z)\propto \big|K_Z\big|.
    \label{eq:DPP}
\end{align}

To help motivate our BO-specific inducing point selection routine, it is worth stressing a link between the DPP formulation above and experimental design. In particular, the arguments of the celebrated paper of \citet{srinivas2009gaussian} can be modified to show that the  MAP $M$-DPP objective of Eq. \ref{eq:DPP} is equivalent to choosing the $M$ inducing points $\mathcal{Z}$ providing the largest gain in information about the unknown function $f$,
\begin{align}
    \mathcal{Z}=\argmax_{Z\subseteq D_n : |Z|=M} \textrm{MI}(\textbf{y}_Z,f),
    \label{eq:maxvar}
\end{align}
where the mutual information $\textrm{MI}(\textbf{y}_Z,f) = H(\textbf{y}_Z)-\mathds{E}\left[H(\textbf{y}_Z|f)\right]$ quantifies the reduction in the differential entropy $H$ of $f$ provided by revealing the  evaluations $\textbf{y}_Z$. 

Maximising Eq. \ref{eq:DPP} corresponds to the \textit{maximum a posteriori} (MAP) estimation of a DPP, which is known to be an NP-hard problem \citep{ko1995exact}, making the use of Eq. \ref{eq:DPP} impractical. 
However, it is possible to obtain a high-performing approximation of the MAP estimate using a greedy algorithm which we refer to as conditional variance reduction (CVR).
In practice, this algorithm can be seen to greedily build a set of inducing points by maximising the posterior predictive variance of a noise-free and zero-mean Gaussian process model $f\sim \mathcal{GP}(0, k)$ conditioned on previously selected observations.
Through incremental updates of Cholesky decompositions, this algorithm returns $M$ points from a set of $N$ candidates with only $O(M^2N)$ complexity, i.e. the same cost as a single fit of the resulting sparse GP model \citep[see ][for detailed description and discussion]{hennig2016exact,chen2018fast,burt2019rates}.

\section{Inducing Point Selection for Bayesian Optimisation Surrogate Models}
Our primary considerations to build a BO-specific routine are a) to allow the focusing of modelling resources into promising areas of the space whilst b) maintaining a sufficiently accurate global model. Accurate modelling in promising areas are necessary to allow the precise identification of the optimum, while a level of global accuracy is necessary to prevent the re-investigation of areas already identified as sub-optimal.

\subsection{Reducing Uncertainty in $f^*$}

We propose choosing a set of inducing points that provide an optimal trade-off between providing information about the whole objective function \textbf{and} providing information about the function's maximum value. This idea is motivated by the empirical success of max-value entropy search acquisition functions \citep{wang2017max, takeno2019multi, moss2020mumbo}, where query points in BO loops are chosen to reduce our current uncertainty in the function's maximal value $f^* = \max f(\textbf{x})$. 

This entropy-based criterion for inducing point placement, henceforth referred to as ENT-DPP (for reasons that will become obvious below), seeks inducing points $Z$ that maximise the following trade-off:
\begin{align}
     C(Z) =\alpha \times\textrm{IG}(\textbf{y}_Z;f^*) +  (1-\alpha) \times \textrm{IG}(\textbf{y}_Z;f),
    \label{eq:MES}
\end{align}
where $\textrm{IG}(\textbf{y}_Z,f^*) = H(f^*)- H(f^*|\textbf{y}_Z)$ quantifies the reduction in uncertainty provided by these inducing points about the objective function's maximal value $f^*$, and $\textrm{IG}(\textbf{y}_Z;f)$ is an in Eq. \ref{eq:maxvar}. 
The trade-off parameter $\alpha \in [0,1]$ controls the balance of resources spent modelling global variations and modelling variations around potential maxima. Note that setting $\alpha=1$ returns the criterion of Eq. \ref{eq:maxvar}. In our experiments we consider an equal trade-off (ie.. $\alpha=0.5$); however, future work will investigate the practical and theoretical effects of other choices. 

\begin{figure}%
\subfloat[K-means]{\includegraphics[height= 0.28\textwidth]{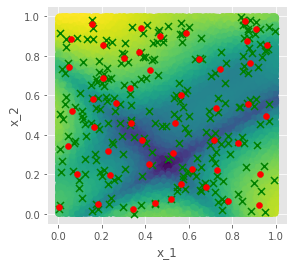}}%
\subfloat[CVR]{\includegraphics[height= 0.28\textwidth]{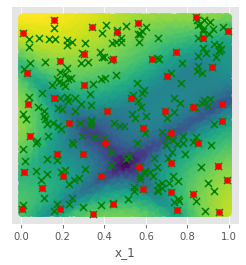}}%
\subfloat[ENT-DPP]{\includegraphics[height= 0.28\textwidth]{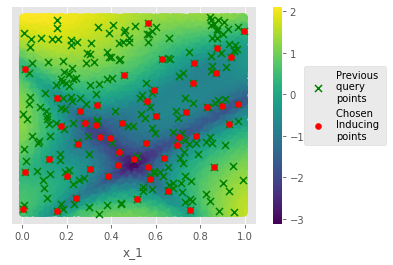}}%
\caption{ The 50 inducing points (red) chosen from 250 candidate evaluations (green) using three different inducing point selection strategies when seeking to minimise the 2d Logarithmic Goldstein-price function (background colour-map). We see that existing approaches that (a) use the centroids from a k-means clustering of the available data or (b) use the CVR strategy provide balanced coverage of the whole search space, whereas our ENT-DPP strategy (c) is able to focus resources modelling resources into promising (blue) areas.
}%
\label{fig::inducing_points}%
\end{figure}

As demonstrated in Figure \ref{fig::inducing_points}, we now have an intuitive inducing point selection strategy well-suited to the demands of BO; however, our criterion is only practically useful if its inducing points can be identified with at most $O(M^2N)$ cost.

\subsection{Efficient (Approximate) Maximisation of the ENT-DPP Criterion}
The lack of closed-form expression for the distribution of a Gaussian process's maximum value $f^*$ renders the calculation of the ENT-DPP criterion (\ref{eq:MES}) challenging. Fortunately, there exists a rich literature of methods for approximately optimising similar information-theoretic quantities \citep{hennig2012entropy,hernandez2014predictive, moss2021gibbon}. 

We follow the ideas of \cite{moss2021gibbon} and use common information-theoretic inequalities to replace our desired criterion (which we will be attempting to maximise) with a simpler lower bound. Specifically, we use the well-known fact that differential entropy reduces under conditioning, i.e. that $H(f^*|\textbf{y}_{Z})\leq H(f^*|y_{\textbf{z}_i})\, \forall i\in \{1,..,M\}$ and write $H(f^*|\textbf{y}_{Z})\leq\frac{1}{M}\sum_{i=1}^M H(f^*|y_{\textbf{z}_i})$. After simple mathematical manipulation, the resulting lower bound for the ENT-DPP criterion can be expressed in the following form:
\begin{align}
    C(Z) \geq \frac{1-\alpha}{2}\log \big| L_{Z}\big|, \label{lowerbound} 
\end{align}
for $L_Z = \textbf{q}_Z^TK_Z\textbf{q}_Z$, where $K_Z$ is the kernel Gram matrix and where  $\textbf{q}_Z$ is the diagonal matrix containing the elements $q_\textbf{z}$ defined such that $(M(1-\alpha) / \alpha)\log q_\textbf{z}  = \textrm{IG}(y_{\textbf{z}};f^*)$. Note that we cannot calculate $\textrm{IG}(y_{\textbf{z}};f^*)$ exactly, so we follow \cite{ru2018fast} and use a moment-matching approximation. By replacing $f^*$ with a Gaussian variable of mean $\mu$ and variance $\sigma^2$ (as extracted from a set of sampled $f^*$), we have  $\textrm{IG}(y_{\textbf{z}};f^*) \approx \frac{\gamma\phi(\gamma)}{2\Phi(\gamma} - \log \Phi(\gamma)$ for $\gamma = \frac{y_{\textbf{z}_i - \mu}}{\sigma}$ (see \cite{wang2017max} for a derivation and strategies for sampling $f^*$).

Just like the criterion described in section \ref{CVR}, maximising the derived lower bound (\ref{lowerbound}) corresponds to the MAP estimate of a DPP, however, with $K_Z$ replaced by $L_Z$.
Hence, we can employ exactly the same greedy algorithm to efficiently build our set of inducing points.

\section{Experimental Results}
\label{exp}
We now provide an empirical evaluation  across high-throughput versions of some popular synthetic optimisation benchmarks using the open-source BO library \texttt{trieste} \citep{Berkeley_Trieste_2022}.

For clarity, all our experiments follow the same setup. We consider an SVGP model with either $M=250$ or $500$ inducing points using either 1) our proposed ENT-DPP method, 2) the CVR of \cite{burt2019rates} (see Section \ref{CVR}), 3) choosing the centroids of a K-means clustering of the data, and 4) choosing inducing points spread uniformly across the search space. When the total number of queried points is less than the desired number of inducing points (e.g. for the first 4 optimisation steps when $M=500$), we use just the $N$ available training points as our inducing point locations. SVGP models are fit using an ADAM optimiser with learning rate $0.1$, using an early stopping criteria with a patience of $50$ and a learning rate halving schedule with a patience of $10$. 

We allocate a total evaluation budget of $N=5,000$ evaluations split across $50$ BO steps in batches of $100$ points, starting from an initial batch of random evaluations. Subsequent batches are collected via the Decoupled Thompson sampling scheme presented in \cite{vakili2021scalable}. We use $100$ random Fourier features to build the samples and maximise each sample using an L-BFGS optimiser starting from the best of a random sample of $1,000$ points. For additional context, BO using an exact GP model is included as a baseline, however, we can report only the first $10$ optimisation steps after which it became prohibitively expensive (i.e. for $N>1,000$).

\begin{figure}%
\subfloat[4d Shekel Function]{\includegraphics[height = 0.23\textwidth]{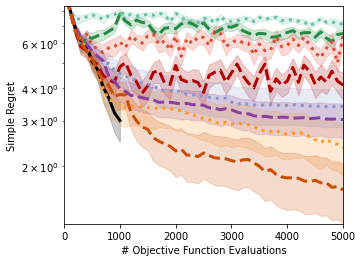}}%
\subfloat[5d Michalewicz Function]{\includegraphics[height= 0.23\textwidth]{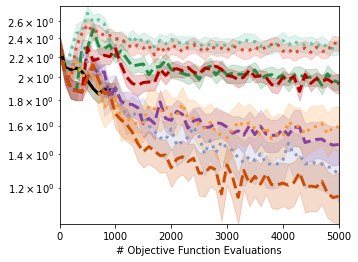}}%
\subfloat[4d Ackley Function]{\includegraphics[height= 0.23\textwidth]{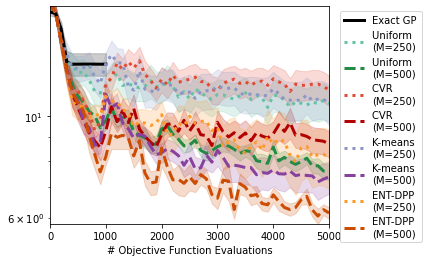}}%
\caption{ Results are averaged over $50$ runs and we report the mean its $95\%$ confidence intervals for the simple regret of the current believed best solution, i.e. the maximiser of the
model mean across previously queried points. Our ENT-DPP is the only method that consistently provides high-performance across the three tasks.
}%
\label{fig::results}%
\end{figure}

Figure \ref{fig::results} demonstrates experimental optimisation performance across the 4d Shekel, 5d Michalewicz and 4d Ackley function, where we have contaminated the evaluations of each with Gaussian noise of variance $0.1$. Unsurprisingly, greater performance is achieved when using larger number of inducing points for all the considered methods, with ENT-DPP providing the lowest regret across all three tasks. Note that, ENT-DPP outperforms all other approaches on the Shekel function even when using just $M=250$ inducing points. Interestingly, the SVGP-based methods are less prone to converging to one of the many local minima of the Ackley function than exact GPs. Future work will investigate exactly why this is the case.


\section{Conclusions and Future Work}
We have proposed ENT-DPP --- the first BO-specific method for selecting the locations for the inducing points of sparse GPs. Although ENT-DPP's approximate maximisation relies on potentially coarse approximations, our experimentation found this new inducing point allocation strategy to enable effective high-throughput BO. In future work we will apply ENT-DPP to real-world  problems where sparse GPs are already being used, e.g. quantile optimisation \citep{torossian2020bayesian} or molecular search \citep{vakili2021scalable}. We will also investigate the applicability of ENT-DPP to other inducing point-based methods like Deep GPs \citep{damianou2013deep}.

\vskip 0.2in
\bibliography{refs}

\begin{thebibliography}{22}
\providecommand{\natexlab}[1]{#1}
\providecommand{\url}[1]{\texttt{#1}}
\expandafter\ifx\csname urlstyle\endcsname\relax
  \providecommand{\doi}[1]{doi: #1}\else
  \providecommand{\doi}{doi: \begingroup \urlstyle{rm}\Url}\fi

\bibitem[Berkeley et~al.(2022)Berkeley, Moss, Artemev, Pascual-Diaz, Granta,
  Stojic, Couckuyt, Qing, Loka, Paleyes, Ober, and
  Picheny]{Berkeley_Trieste_2022}
Joel Berkeley, Henry~B. Moss, Artem Artemev, Sergio Pascual-Diaz, Uri Granta,
  Hrvoje Stojic, Ivo Couckuyt, Jixiang Qing, Nasrulloh Loka, Andrei Paleyes,
  Sebastian~W. Ober, and Victor Picheny.
\newblock {Trieste}.
\newblock \emph{https://github.com/secondmind-labs/trieste}, 2022.

\bibitem[Burt et~al.(2019)Burt, Rasmussen, and Van Der~Wilk]{burt2019rates}
David Burt, Carl~Edward Rasmussen, and Mark Van Der~Wilk.
\newblock Rates of convergence for sparse variational gaussian process
  regression.
\newblock In \emph{International Conference on Machine Learning}, pages
  862--871. PMLR, 2019.

\bibitem[Chen et~al.(2018)Chen, Zhang, and Zhou]{chen2018fast}
Laming Chen, Guoxin Zhang, and Hanning Zhou.
\newblock Fast greedy map inference for determinantal point process to improve
  recommendation diversity.
\newblock In \emph{Proceedings of the 32nd International Conference on Neural
  Information Processing Systems}, pages 5627--5638, 2018.

\bibitem[Damianou and Lawrence(2013)]{damianou2013deep}
Andreas Damianou and Neil~D Lawrence.
\newblock Deep gaussian processes.
\newblock In \emph{Artificial intelligence and statistics}, pages 207--215.
  PMLR, 2013.

\bibitem[Hennig and Garnett(2016)]{hennig2016exact}
Philipp Hennig and Roman Garnett.
\newblock Exact sampling from determinantal point processes.
\newblock \emph{arXiv preprint arXiv:1609.06840}, 2016.

\bibitem[Hennig and Schuler(2012)]{hennig2012entropy}
Philipp Hennig and Christian~J Schuler.
\newblock Entropy search for information-efficient global optimization.
\newblock \emph{Journal of Machine Learning Research}, 13\penalty0 (6), 2012.

\bibitem[Hensman et~al.(2013)Hensman, Fusi, and Lawrence]{hensman2013gaussian}
James Hensman, Nicolo Fusi, and Neil~D Lawrence.
\newblock Gaussian processes for big data.
\newblock \emph{arXiv preprint arXiv:1309.6835}, 2013.

\bibitem[Hern{\'a}ndez-Lobato et~al.(2014)Hern{\'a}ndez-Lobato, Hoffman, and
  Ghahramani]{hernandez2014predictive}
Jos{\'e}~Miguel Hern{\'a}ndez-Lobato, Matthew~W Hoffman, and Zoubin Ghahramani.
\newblock Predictive entropy search for efficient global optimization of
  black-box functions.
\newblock \emph{arXiv preprint arXiv:1406.2541}, 2014.

\bibitem[Ko et~al.(1995)Ko, Lee, and Queyranne]{ko1995exact}
Chun-Wa Ko, Jon Lee, and Maurice Queyranne.
\newblock An exact algorithm for maximum entropy sampling.
\newblock \emph{Operations Research}, 43\penalty0 (4):\penalty0 684--691, 1995.

\bibitem[Kulesza and Taskar(2012)]{kulesza2012determinantal}
Alex Kulesza and Ben Taskar.
\newblock Determinantal point processes for machine learning.
\newblock \emph{arXiv preprint arXiv:1207.6083}, 2012.

\bibitem[Moss et~al.(2020)Moss, Leslie, and Rayson]{moss2020mumbo}
Henry~B Moss, David~S Leslie, and Paul Rayson.
\newblock Mumbo: Multi-task max-value bayesian optimization.
\newblock \emph{arXiv preprint arXiv:2006.12093}, 2020.

\bibitem[Moss et~al.(2021)Moss, Leslie, Gonzalez, and Rayson]{moss2021gibbon}
Henry~B Moss, David~S Leslie, Javier Gonzalez, and Paul Rayson.
\newblock Gibbon: General-purpose information-based bayesian optimisation.
\newblock \emph{arXiv preprint arXiv:2102.03324}, 2021.

\bibitem[Rasmussen(2004)]{rasmussen2003gaussian}
Carl~Edward Rasmussen.
\newblock {G}aussian processes in machine learning.
\newblock In \emph{Advanced Lectures on Machine Learning}, pages 63--71.
  Springer, 2004.

\bibitem[Ru et~al.(2018)Ru, Osborne, McLeod, and Granziol]{ru2018fast}
Binxin Ru, Michael~A Osborne, Mark McLeod, and Diego Granziol.
\newblock Fast information-theoretic bayesian optimisation.
\newblock In \emph{International Conference on Machine Learning}, pages
  4384--4392. PMLR, 2018.

\bibitem[Shahriari et~al.(2016)Shahriari, Swersky, Wang, Adams, and
  De~Freitas]{shahriari2016taking}
Bobak Shahriari, Kevin Swersky, Ziyu Wang, Ryan~P Adams, and Nando De~Freitas.
\newblock Taking the human out of the loop: A review of {B}ayesian
  optimization.
\newblock \emph{Proceedings of the IEEE}, 2016.

\bibitem[Srinivas et~al.(2009)Srinivas, Krause, Kakade, and
  Seeger]{srinivas2009gaussian}
Niranjan Srinivas, Andreas Krause, Sham~M Kakade, and Matthias Seeger.
\newblock Gaussian process optimization in the bandit setting: No regret and
  experimental design.
\newblock \emph{arXiv preprint arXiv:0912.3995}, 2009.

\bibitem[Takeno et~al.(2019)Takeno, Fukuoka, Tsukada, Koyama, Shiga, Takeuchi,
  and Karasuyama]{takeno2019multi}
Shion Takeno, Hitoshi Fukuoka, Yuhki Tsukada, Toshiyuki Koyama, Motoki Shiga,
  Ichiro Takeuchi, and Masayuki Karasuyama.
\newblock Multi-fidelity bayesian optimization with max-value entropy search.
\newblock \emph{arXiv preprint arXiv:1901.08275}, 2019.

\bibitem[Titsias(2009)]{titsias2009variational}
Michalis Titsias.
\newblock Variational learning of inducing variables in sparse gaussian
  processes.
\newblock In \emph{Artificial intelligence and statistics}, pages 567--574.
  PMLR, 2009.

\bibitem[Torossian et~al.(2020)Torossian, Picheny, and
  Durrande]{torossian2020bayesian}
L{\'e}onard Torossian, Victor Picheny, and Nicolas Durrande.
\newblock Bayesian quantile and expectile optimisation.
\newblock \emph{arXiv preprint arXiv:2001.04833}, 2020.

\bibitem[Vakili et~al.(2021)Vakili, Moss, Artemev, Dutordoir, and
  Picheny]{vakili2021scalable}
Sattar Vakili, Henry Moss, Artem Artemev, Vincent Dutordoir, and Victor
  Picheny.
\newblock Scalable thompson sampling using sparse gaussian process models.
\newblock \emph{Advances in Neural Information Processing Systems}, 34, 2021.

\bibitem[Wang and Jegelka(2017)]{wang2017max}
Zi~Wang and Stefanie Jegelka.
\newblock Max-value entropy search for efficient bayesian optimization.
\newblock In \emph{International Conference on Machine Learning}, pages
  3627--3635. PMLR, 2017.

\bibitem[Wilson et~al.(2020)Wilson, Borovitskiy, Terenin, Mostowsky, and
  Deisenroth]{wilson2020efficiently}
James Wilson, Viacheslav Borovitskiy, Alexander Terenin, Peter Mostowsky, and
  Marc Deisenroth.
\newblock Efficiently sampling functions from gaussian process posteriors.
\newblock In \emph{International Conference on Machine Learning}, pages
  10292--10302. PMLR, 2020.

\end{thebibliography}

\end{document}